\documentclass[letterpaper, 10 pt, conference]{ieeeconf}

\IEEEoverridecommandlockouts
\usepackage{cite}
\usepackage{amsmath,amssymb,amsfonts}
\usepackage{algorithmic}
\usepackage{graphicx}
\usepackage{textcomp}
\usepackage{xcolor}
\usepackage{hyperref}
\usepackage{booktabs}
\usepackage{multirow}
\usepackage{subcaption}
\usepackage{float}
\usepackage[utf8]{inputenc}
\usepackage[T1]{fontenc}

\title{\LARGE \bf
Persistent Metric-Semantic Maritime Scene Memory from Open-Vocabulary 3D Scene Graphs
}

\author{Kim Alexander Christensen$^{1}$, Dina Zimmermann$^{1}$,
Martin Steinert$^{1,\dagger}$ and Eleni Kelasidi$^{1,\dagger}$%
\thanks{$^{1}$All authors are with the Department of Mechanical and
Industrial Engineering, Norwegian University of Science and Technology
(NTNU), Norway. Corresponding author:
{\tt\small kim.a.christensen@ntnu.no}}%
\thanks{$^{\dagger}$This work was co-advised.}%
}

\begin{document}

\maketitle
\thispagestyle{empty}
\pagestyle{empty}

\begin{abstract}

Three-dimensional scene graphs (3DSGs) can provide robots with persistent metric-semantic representations for spatial reasoning, but the existing methods have primarily been developed for indoor and terrestrial environments. In this paper we investigate how a representative outdoor open-vocabulary method transfers to maritime environments and which adaptations are needed. Our experiments show that zero-shot representative outdoor 3DSG descriptions agree with our labeled ground truth for a majority of geometry-supported objects on our data, but the resulting map exhibits considerable physical-entity fragmentation, and its terrestrial higher-level assumptions don't capture maritime spatial organization needs. We therefore keep the underlying mapping method and post-process its accumulated evidence. We show a metric-semantic consolidation method followed by geometry-grounded multi-view vision-language model (VLM) reasoning that reconciles remaining fragments into physical entities and groups entities through observations into local places. In parallel, we construct maritime macro regions using map priors and onboard water and structure observations, and associate them with the local places. The resulting representation reduces physical-entity fragmentation beyond both the baseline and our geometry-semantic consolidation component alone, supports correspondence of fixed local places across repeated maritime traversals, and improves contextual retrieval on our reviewed queries over tested baselines. These results show how an outdoor open-vocabulary map can be adapted into persistent maritime scene memory with our methods for object/place identity and spatial abstractions.

\end{abstract}

\section{INTRODUCTION}
\label{sec:introduction}

Autonomous surface vessels (ASVs) are being developed for transportation, inspection, monitoring, and other repeated operations in harbors, inland waterways, and coastal areas~\cite{Brekke2022,blanke2024greenhopper,akbar2021economic}. Such operations require more than just obstacle tracking and detection in the moment. A vessel revisiting the same area should retain a memory of persistent objects, distinguish them from temporary occupants, and reason about semantic observations in their spatial context. This motivates a metric-semantic scene memory of the maritime environment.

\begin{figure}[!t]
\centering
\includegraphics[width=0.95\linewidth]{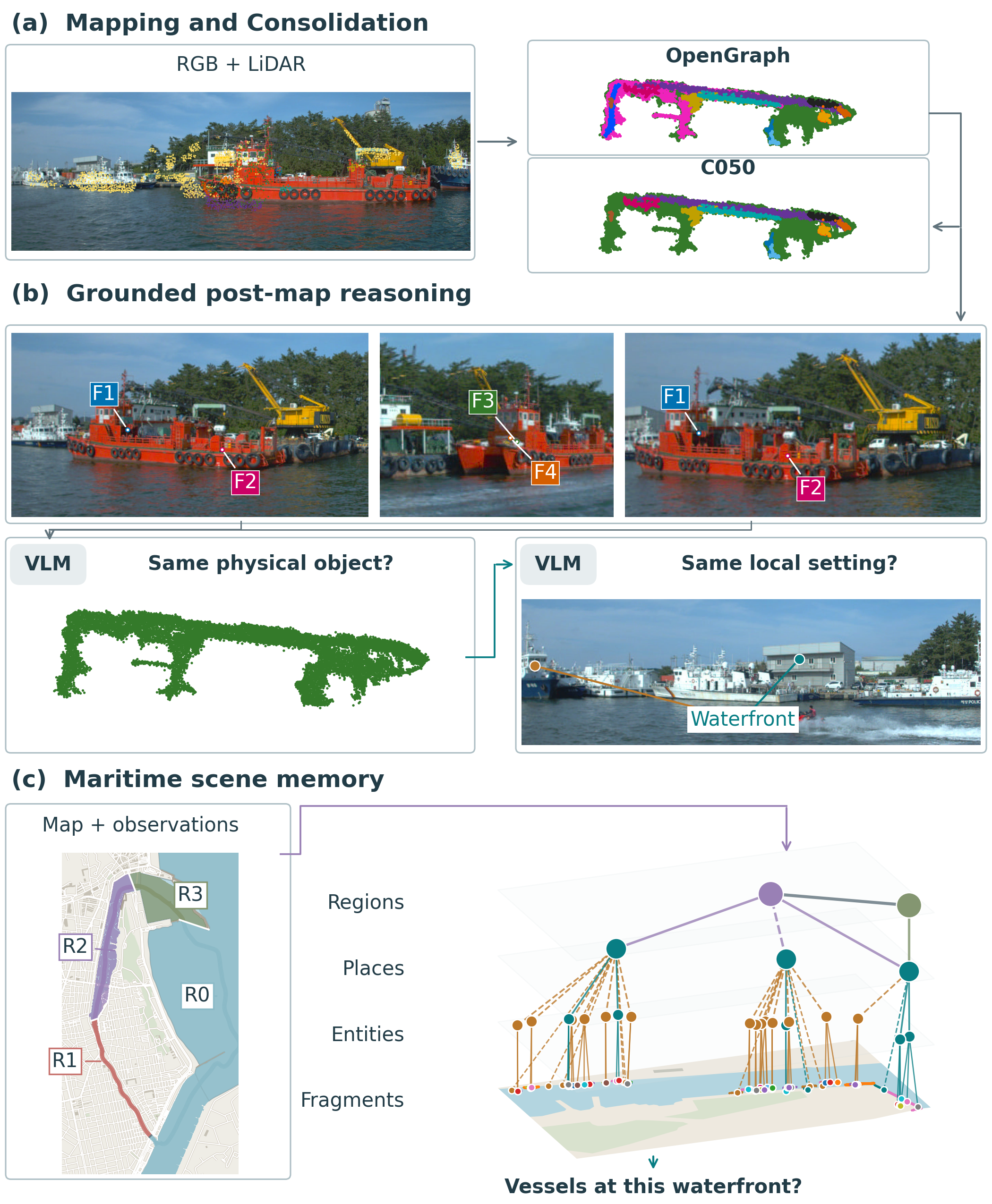}
\caption{Overview of the maritime scene-memory pipeline. It first accumulates persistent map fragments and their observation histories, followed by a conservative geometric-semantic consolidation. Selected geometrically grounded multi-view observations are then processed by a vision-language model (VLM) at two abstraction levels: physical-entity reconciliation decides which fragments belong to the same object, while place formation groups persistent entities and observations into meaningful local places. In parallel, coarse maritime regions are derived from geographic topology and onboard water/structure evidence. Local places are associated with these regions and corresponding places are related across repeated traversals, forming a persistent scene memory for contextual retrieval.}
\label{fig:method}
\vspace{-15pt}
\end{figure}

Three-dimensional scene graphs (3DSGs) can provide such representations by connecting metric geometry with semantic entities, relations, and higher-level spatial abstractions~\cite{armeni20193d,rosinol20203d,rosinol2021kimera,hughes2022hydra}. Recent open-vocabulary systems extend this capability beyond the previously fixed semantic taxonomies~\cite{gu2024conceptgraphs,werby2024hovsg,deng2024opengraph}. However, 3DSG hierarchies are strongly domain dependent. Indoor systems are concerned with rooms and floors, while outdoor systems use abstractions such as roads, lanes, terrain, or task-dependent regions~\cite{greve2024collaborative,strader2024indoor,samuelson2025terra,tonetti2026priorsg}. An ASV instead operates through large scale open areas like canals, harbors and open water. Existing outdoor open-vocabulary 3DSGs do not establish which parts of their representation transfer to maritime environments, and which parts require domain-specific adaptations.

We study this transfer using OpenGraph~\cite{deng2024opengraph}, a representative hierarchical open-vocabulary RGB--LiDAR 3DSG designed for large-scale outdoor mapping, as a baseline. We find that for discrete maritime objects with good geometric map support, accumulated OpenGraph descriptions agree with the ground-truth class in 99.1\% of cases, while extended and partially hidden waterfront structures are less reliable. Despite this transfer, the resulting mapped object representation shows considerable fragmentation, while its terrestrial higher-level organization assumptions does not provide the spatial structure required for maritime scene understanding. We therefore use the released perception and incremental mapping substrate while contributing a maritime-adapted method for fragment consolidation followed by grounded multi-view vision-language reasoning at the physical-entity and local-place levels. Higher level maritime macro regions are constructed separately from geographic topology and onboard water/structure evidence. Our proposed method is summarized in Fig.~\ref{fig:method}. 

The complete adaptation reduces physical-entity fragmentation while preserving vessel recovery, reducing nodes per recovered vessel from $1.957$ to $1.298$ and fragmented vessels from $25$ to $11$ in our evaluation. We show that independently constructed local places correspond across traversals despite differences in reconstruction and different object occupants, with 12 reciprocal correspondences across ten evaluated geographic contexts. The resulting contextual representation also improves paired-survey retrieval F1 from $0.378$ to $0.810$ compared to a local metric baseline.

Our contributions are as follows:

\begin{itemize}

    \item \textbf{A layer-wise maritime transfer evaluation of an outdoor open-vocabulary 3DSG.} We show that semantic descriptions transfer well for geometry-supported discrete objects, while mapped physical-entity consistency and higher-level spatial organization require adaptation.
    
    \item \textbf{A grounded maritime scene-memory construction.} We keep the underlying open-vocabulary perception and incremental mapping pipeline, but adapt its accumulated representation through maritime-specific consolidation and grounded multi-view vision-language model (VLM) reasoning to form physical entities and local places. Higher-level maritime context is constructed separately by combining geographic topology with onboard water and structure observations. 
    
    \item \textbf{An evaluation of persistence and contextual use across repeated maritime surveys.} We evaluate physical-entity consistency, cross-traversal local-place correspondence, macro-region repeatability, and contextual retrieval compared to simpler spatial baselines.

\end{itemize}

\section{RELATED WORK}
\label{sec:related_work}

3DSGs combine metric grounding with semantic entities, relations, and/or hierarchical spatial abstractions~\cite{armeni20193d,rosinol20203d,rosinol2021kimera,hughes2022hydra}. Further extensions include methods for task dependent abstractions, dynamic environments, and long-term memory~\cite{schmid2024khronos,maggio2024clio,gorlo2026daaam}. In recent years, foundation models have enabled open vocabulary objects and relations~\cite{gu2024conceptgraphs,koch2024open3dsg,werby2024hovsg}, where OpenGraph is a relevant example as a hierarchical open-vocabulary camera-LiDAR 3DSG designed for large-scale outdoor environments~\cite{deng2024opengraph}. Incremental 3DSGs relevant for robotics typically lift frame-level observations into 3D and then associate them across viewpoints using geometric and semantic cues, which can yield duplicated, fragmented, merged, or missing object nodes because of partial visibility, ambiguous boundaries and geometry~\cite{gu2024conceptgraphs,deng2024opengraph,hughes2022hydra,rotondi2026scenegraphs}. These problems are often exacerbated outdoors~\cite{lim2026hydra++}.
To combat this, ThinkGraphs selects geometrically and semantically plausible duplicate tracks, finds informative views, and uses a VLM to verify merges \cite{bickici2026thinkgraphs}. DAAAM~\cite{gorlo2026daaam} similarly performs later semantic annotation, fragment reconciliation, and hierarchical region construction, while relationship-aware
3DSGs prompt VLMs for hierarchical relationships~\cite{gassol2026relationship}. 

Higher-level 3DSG graph structure is environment dependent, with indoor systems typically organizing objects through places, rooms, floors and buildings~\cite{rosinol20203d,rosinol2021kimera,hughes2022hydra,werby2024hovsg}, while outdoor systems need different abstractions~\cite{strader2024indoor,rotondi2026scenegraphs,greve2024collaborative}. Car-based approaches such as OpenGraph, CURB-SG, and CURB-OSG take advantage of trajectories, roads, lanes, intersections and curbs to form their graph~\cite{deng2024opengraph,greve2024collaborative,steinke2025collaborative}. Other approaches organize outdoor spaces using semantic composition~\cite{strader2024indoor}, terrain and traversability
\cite{samuelson2025towards,samuelson2025terra}, or multiscale object parts~\cite{nyffeler2025hierarchical}. The high-level abstractions can also depend on the task at hand~\cite{maggio2024clio}, prior knowledge~\cite{tonetti2026priorsg}, abstractions derived from large language models (LLMs) or VLMs~\cite{berg2022using,hao2026ovipsg},  or specific missions~\cite{viswanathan2026xflie}. No single outdoor hierarchy is universal, and existing formulations are largely organized around terrestrial cues like roads and terrain or task priors.

Maritime surface robotics already has domain-specific methods and benchmarks on visual obstacle and water perception~\cite{bovcon2021mods,zust2023lars}, while SGK-Net~\cite{yang2024sgknet} applies graph-structured reasoning to maritime 2D navigational images. Vision-language models have also been used in maritime autonomy for open vocabulary scene understanding~\cite{christensen2026foundation}. These methods establish maritime semantic and relational perception, but don't construct a persistent semantic 3DSG. LiDAR SLAM has also been demonstrated for maritime mapping and localization in coastal areas~\cite{sawada_mapping_2023,engstrom_lidar-only_2022}. Further, there is existing work in persistent geographic structure representing inland waterways through semantic world models based on chart-derived static objects and runtime context~\cite{vanbaelen2022dynamic}, as well as Oceanscape which constructs coastal navigation graphs from geospatial information~\cite{fagerhaug2025oceanscape}. Existing near-shore mapping works also combines onboard perception with prior map geometry~\cite{dalhaug2025near}, and coastline geometry has been used as a structural constraint for ASV localization~\cite{benham2026coastline}, the commonality is that map priors are used together with onboard observations, rather than reconstructing large-scale maritime environments from sensing alone. Many terrestrial surfaces can be represented directly by LiDAR or terrain semantics. Water provides negligible LiDAR surface returns~\cite{chung2023pohang}, which is why visual water evidence can provide additional useful spatial information for ASV operations~\cite{bovcon2021mods,zust2023lars}. In our work, we represent large scale geographic information with map topology, while onboard water/structure observations ground the currently observed scene.

Despite existing work in 3DSGs and maritime autonomy, persistent maritime 3DSGs that organize physical entities, local places, and geographic context across repeated surveys remain largely unexplored.

\section{MARITIME SCENE-MEMORY ADAPTATION}
\label{sec:method}

As will be shown, direct transfer of OpenGraph leaves two main representation limitations: fragmented physical identities and no appropriate higher-level maritime spatial organization. We therefore keep the released perception and incremental mapping substrate while adapting the representation at the levels where transfer is insufficient. Our conservative consolidation and grounded multi-view reasoning methods predict physical entities and local places, while geographic topology together with onboard water and structure observations independently provides region-level maritime context. Fixed local place infrastructure is subsequently used to relate corresponding places across repeated traversals.

\subsection{Problem Formulation}
\label{sec:problem}

Given posed and calibrated RGB--LiDAR observations from traversal $t$, an outdoor open-vocabulary mapper produces a set of persistent map fragments $\mathcal{F}^{t}$. Each fragment contains reconstructed geometry, semantic attributes, and its observation history, but a fragment does not necessarily correspond exactly with a physical object.

Our objective is to construct a maritime scene representation with increasingly abstract spatial entities,

\begin{equation}
\mathcal{F}^{t}
\rightarrow
\mathcal{E}^{t}
\rightarrow
\mathcal{P}^{t},
\label{eq:hierarchy}
\end{equation}

where $\mathcal{E}^{t}$ denotes physical entities and $\mathcal{P}^{t}$ local places. In parallel, maritime macro regions $\mathcal{R}^{t}$ are constructed in the world frame from map topology and traversal-dependent water and structure evidence. Local places are then spatially associated with these regions. Uncertain fragments or entities can remain unassigned.

For repeated surveys, a place is considered persistent when it is supported by the same underlying fixed infrastructure, even if only part of that structure is observed and its movable occupants differ between traversals. Thus, stable infrastructure defines place identity, while vessels and other movable objects describe traversal-dependent scene state. We focus on static structures and objects that are stationary during a traversal, including moored vessels. Dynamic-object reconstruction and tracking are outside the scope of this work.

\subsection{Outdoor Mapping and Conservative Consolidation}
\label{sec:method_mapping}

We use the released OpenGraph~\cite{deng2024opengraph} perception and mapping pipeline through dataset-specific adapters. Open-vocabulary image observations are lifted using LiDAR and incrementally associated into persistent fragments, retaining their geometry, semantic attributes, and observation histories. Details can be found in~\cite{deng2024opengraph}. We keep the corresponding OpenGraph configuration fixed within each comparison.

The first adaptation is a conservative consolidation stage C050 to remove clear duplicate fragments before higher-level reasoning. For a smaller source point cloud $X$ and candidate destination point cloud $Y$, define directed geometric support

\begin{equation}
o_d(X,Y)=
\frac{|\{x\in X:\min_{y\in Y}\|x-y\|\leq d\}|}{|X|}.
\end{equation}

where $d$ is the Euclidean support tolerance.

A merge is considered when the fragments have a 3D bounding-box gap of at most $0.5$\,m, their semantic-feature cosine similarity is at least $0.8$, $o_{0.5}(X,Y)\geq0.7$ and $o_{0.5}(Y,X)\geq0.1$. For co-observed fragments, the source-support requirement is increased to $0.9$. We additionally reject merges that expand either horizontal extent beyond $1.2$ times the larger input extent plus $0.25$\,m. Accepted merges subsequently use the released OpenGraph fusion operation while preserving fragment lineage for the later multi-view reasoning experiments.

\subsection{Grounded Multi-View Physical-Entity Reasoning}
\label{sec:method_entities}

The previous stage removes clear duplicates, but one physical object can still be represented by several fragments. We therefore post-process the accumulated map by having a VLM reason over selected observations from its history. Candidate fragment groups contain a maximum of 18 elements, with a sampled support gap of at most $8$\,m and center separation of at most $45$\,m. The candidate geometry is projected back into recorded RGB observations, and views are kept when their mapped support is geometrically visible and consistent with the current LiDAR returns. Numbered overlays identify the candidate fragments in each view.

For physical-entity reconciliation, the VLM (GPT-6 Astra~\cite{openai2026astra}) receives three complementary views together with geometry diagnostics for all candidate fragments. It first proposes which fragments belong to the same physical object, followed by a contrastive verification using the same evidence and proposed group. The verification considers whether the fragments instead represent separate physical objects. Only supported and geometrically consistent groups are joined into physical entities. Each accepted group forms a parent entity while retaining its constituent fragment IDs and lineage. Unsupported or conflicting groups remain separate. The essence of this method is that the VLM reasons over recorded observations geometrically grounded to candidate map fragments rather than over an RGB image or semantic descriptions alone.

\subsection{Grounded Local-Place Reasoning}
\label{sec:method_places}

At the next abstraction level, place reasoning uses broader multi-view context to determine which entities and observations then form one persistent local setting. The candidates are shown in overlapping spatial windows rather than proposed place boundaries. The VLM is asked which fixed setting remains when current vessels and activity are removed, which leads to stable infrastructure anchors giving the place an identity while movable objects characterize its current state but do not connect place proposals in isolation. This reasoning is performed independently for each traversal.

The place reasoning starts from six views, supplemented when necessary with targeted RGB image additions for qualifying structural anchors not represented in the initial selection. The supported proposals are reconciled through shared, uncontested stable anchors, meaning that the same mapped infrastructure elements support the proposals being joined. Conflicting assignments continue to stay separate. A place does not need to consist only of already reconciled physical entities. Instead, locally grounded unresolved observations can also contribute to its support.

\subsection{Maritime Macro-Regions}
\label{sec:method_regions}

At a higher scale, we independently construct maritime regions in~$\mathcal{R}^{t}$ from geographic map priors and onboard water and structure observations. Since LiDAR returns are negligible on water, we project camera-observed water masks from the released image perception onto a fixed nominal horizontal water plane using calibrated cameras and supplied poses to provide positive water evidence in the world frame. LiDAR observations provide complementary structure and obstacle evidence.

Let $W_o$ and $W_c^{t}$ denote map-prior and camera-observed water, and $B_o$ and $B_s^{t}$ mapped and observed structural barriers. The represented water domain is

\begin{equation}
W^{t}=(W_o\cup W_c^{t})\setminus(B_o\cup B_s^{t}).
\end{equation}

The onboard evidence is accumulated across observations and represented on a $2$\,m spatial grid, with at most one water vote and one structure vote per keyframe and cell. For accumulated water and structure counts $w$ and $s$, camera-observed water requires $w\geq3$ and $w/(w+s)\geq0.8$. Cells with $s\geq3$ outside the camera-observed water form structural barriers.

Mapped land remain a hard constraint on the large-scale topology, while onboard evidence grounds the water and structure observed during traversal $t$. We estimate local water widths at $8$\,m-spaced probes using $36$ directions out to $300$\,m. Candidate cuts require at least $75\%$ supported water and barrier contact within $8$\,m of their endpoints. Widening candidates additionally require a width ratio of at least $1.8$ and a difference of at least $30$\,m. Map-mouth separators are prioritized, with $250$\,m separator suppression, $2$\,m separator buffers and $4$\,m connectivity closing. Components with at least $20{,}000~\mathrm{m}^2$ of represented water form macro regions. 

Place--region associations use the nearest assigned region cell within $10$\,m of fixed support, retaining weak or multiple associations and the unassigned fraction.

\subsection{Cross-Traversal Place Correspondence}
\label{sec:method_correspondence}

Next, we want to determine if independently constructed places represent the same persistent setting across repeated completed surveys. We measure correspondence based on shared fixed physical support. Movable occupants, semantic descriptions, and macro-region IDs provide no positive correspondence evidence. Fixed place support is represented using $0.5$\,m voxelized geometry from the stable anchors, excluding disputed memberships and movable objects.

At $3$\,m distance, candidate correspondences require at least 50 close support voxels on each side and at least 20\% directed coverage in one direction. Candidates are ranked using the harmonic mean of directed coverage. A reciprocal correspondence additionally requires at least 20\% coverage in both directions, mutual best rank, and a score margin of at least 0.10 over the next candidate on each side. The comparison assumes both traversals are expressed in a common world frame.

\section{EXPERIMENTAL SETUP}
\label{sec:experiments}

\subsection{Datasets and Evaluation Protocol}
\label{sec:datasets}

We use two main sources of data. The first is the single-target MAPPING sequence from the Maritime Urban Tracking dataset (MUT)~\cite{dalhaug2026maritime}. The sequence contains synchronized LiDAR, monocular RGB image frames, and RTK GNSS-based INS poses. We process every tenth LiDAR scan, giving 967 selected keyframes.

We manually annotate 222 object footprints, comprising 139 discrete objects (including 92 vessels) and 83 extended structures. We split the map across the canal and get 42 north-bank vessels for development and 50 south-bank vessels for evaluation. Ground-truth identities are never provided to the mapping or reconciliation methods.

The second source of data is the Pohang Canal Dataset~\cite{chung2023pohang}, where we use repeated traversals P00 and P03. Each traversal covers approximately 7.5 km, with 2,008 and 2,045 accepted 1 Hz keyframes, respectively. The traversals use the same local coordinate system and origin. Entity and place construction use the forward stereo-left camera, four horizontal omni cameras, and the front LiDAR. The macro-region water layer additionally uses the remaining horizontal omni camera. Geographic context is provided by an OpenStreetMap~\cite{openstreetmap} (OSM) prior.

We organize the evaluation around four questions:

\begin{enumerate}
    \item[\textbf{RQ1}] Which parts of an outdoor open-vocabulary 3DSG transfer directly to maritime scenes and which do not?
    \item[\textbf{RQ2}] Can improved consolidation and grounded multi-view VLM reasoning improve physical-entity consistency?
    \item[\textbf{RQ3}] Can the adapted representation form meaningful higher-level maritime macro regions and local places that persist across repeated traversals?
    \item[\textbf{RQ4}] Does the resulting hierarchy improve contextual retrieval across single and repeated surveys compared with simpler spatial representations?
\end{enumerate}


\subsection{Evaluation}
\label{sec:metrics}

\textbf{RQ1.}
We use the OpenGraph baseline B0 to evaluate what transfers directly to the maritime domain. On the MUT dataset, we associate mapped records with the annotations using the geometric support criterion described in the method and normalize their accumulated text to a fixed simple ontology using an LLM text-only normalizer without access to ground-truth labels. We report geometric coverage and semantic agreement among supported objects, separately for discrete objects and extended structures. We also evaluate physical-entity recovery and fragmentation against the vessel annotations. The higher-level transfer is evaluated qualitatively by whether the existing outdoor representation provides meaningful spatial organization for the maritime scenes considered here.

\textbf{RQ2.}
We compare OpenGraph baseline B0, our conservative consolidation C050, and C050+VLM post-processing. A node supports the ground-truth reference vessel when at least 50 entity points lie within $0.25$\,m of its footprint and either 10\% of the node or 200 points satisfy this condition. We report recovery, nodes per recovered vessel, fragmented vessels, and multi-vessel merge evidence, with sensitivity across three support criteria and $0-1$\,m tolerances. Small and uncertain objects were not labeled, so we don't report whole-scene precision.

\textbf{RQ3.}
For macro regions, we compare onboard water/structure, the same evidence with radar, OSM alone, and OSM plus onboard evidence. We measure cross-traversal agreement with ARI on common camera-water cells. We compare against four environmental areas digitized from the dataset authors' illustration~\cite{chung2023pohang} and manually registered using five landmarks. We report the equal-weighted mean IoU over the four one-to-one-matched reference areas on assigned reference-covered water. This provides an approximate reference for geographic plausibility. 

Local semantic place persistence is evaluated on ten matched contexts by independently constructing P00 and P03 and reporting reciprocal, partial, unmatched, and conflicting correspondences together with fixed-support overlap. Without exhaustive ground-truth place boundaries, this measures repeatability rather than boundary accuracy.

\textbf{RQ4.}
We compare flat entity text, lexical- and supplied-location radius retrieval, our proposed hierarchy, and an equivalent flat representation of the same relations. All methods receive the same structured query material, and language parsing is not evaluated. The radius baselines use $50$\,m neighborhoods around either a text-selected (BM25) anchor or a supplied representative place location. For paired queries, the supplied location is the mean anchor-entity center in P00, carried unchanged into P03. The hierarchy retrieves place members and follows stored place correspondences for paired-survey queries. The primary evaluation contains 20 reviewed single-survey and four paired-survey queries. We additionally evaluate 18 macro-context queries comparing direct region assignment, fixed-radius geometry, and hierarchical place--region associations. These combine reviewed semantics with fixed spatial reference rules. We report mean per-query F1 and exact answer sets separately for the two primary cohorts. 

\section{RESULTS}
\label{sec:results}

\subsection{RQ1: Zero-Shot Maritime Transfer}
\label{sec:results_transfer}

The obtained results show that OpenGraph ends up producing useful open-vocabulary object-level observations of the maritime scene without maritime-specific adaptations. On the MUT dataset, 112/139 discrete objects (80.6\%) have geometric support, and accumulated map text agrees with the ground-truth class on 111/112 of these (99.1\%). Extended structures are more difficult, with 46/83 (55.4\%) supported and 15/46 (32.6\%) agreeing with the overall class (these are often partially hidden behind the objects).

However, the persistent object representation is considerably fragmented as seen in the examples in Fig.~\ref{fig:method} and~\ref{fig:entities} and Table~\ref{tab:entities}. Checked against the MUT vessel annotations, B0 recovers 36/42 north-bank and 46/50 south-bank vessels under our primary support criterion, but these are represented by 2.306 and 1.957 nodes per recovered vessel, respectively. Of the recovered vessels, 20/36 in the north and 25/46 in the south are represented by several nodes.

The higher-level OpenGraph representation is organized around terrestrial road and lane structure and does not directly provide appropriate spatial organization for the maritime domain. Thus, the object-level observations transfer usefully, while physical-entity consistency and higher-level spatial organization require adaptation. We therefore keep the perception and object-mapping substrate and adapt the physical entity and higher-level representation in the following experiments.


\subsection{RQ2: Physical-Entity Reconciliation}
\label{sec:results_entities}

\begin{figure}[tbp]
    \centering
    \includegraphics[width=\linewidth]{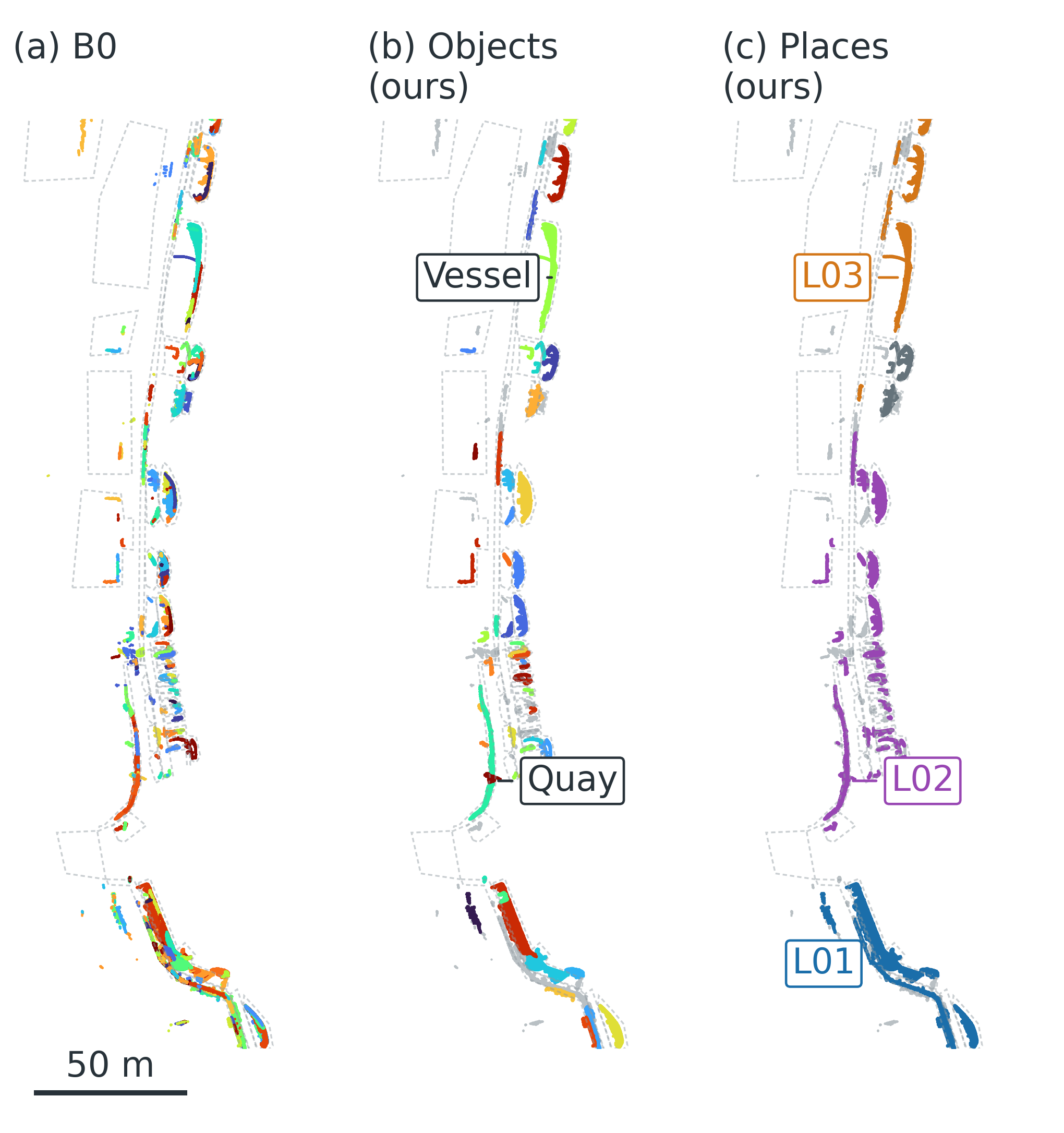}
    \caption{Representative view of the MUT north bank showing (a) B0 fragmentation, (b) C050+VLM objects, and (c) local places.}
    \label{fig:entities}
    \vspace{-20pt}
\end{figure}

Table~\ref{tab:entities} compares OpenGraph B0, our C050 consolidation, and our multi-view VLM reconciliation, while examples are shown in Fig.~\ref{fig:method} and~\ref{fig:entities}. C050 reduces fragmentation substantially while preserving the primary recovery of the baseline. The VLM stage further reduces nodes per recovered vessel from 1.861 to 1.222 on the north bank set and from 1.413 to 1.298 on the south bank set. The number of fragmented vessels decreases from 17 to 8 and from 15 to 11, respectively. Figure~\ref{fig:entities} illustrates object consolidation and place formation on the MUT north bank.

\begin{table}[t]
\centering
\caption{Physical-entity evaluation on MUT dataset. Each bank reports recovered vessels, nodes per recovered vessel, and fragmented recovered vessels.}
\label{tab:entities}
\small
\setlength{\tabcolsep}{3pt}
\begin{tabular}{lccc}
\hline
Method & Recovered & Nodes / rec. & Fragmented \\
\hline
\multicolumn{4}{c}{North development (/42)} \\
OpenGraph B0 & 36 & 2.306 & 20 \\
C050 & 36 & 1.861 & 17 \\
C050 + VLM & 36 & \textbf{1.222} & \textbf{8} \\
\hline
\multicolumn{4}{c}{South evaluation (/50)} \\
OpenGraph B0 & 46 & 1.957 & 25 \\
C050 & 46 & 1.413 & 15 \\
C050 + VLM & \textbf{47} & \textbf{1.298} & \textbf{11} \\
\hline
\end{tabular}
\vspace{-15pt}
\end{table}

Relative to B0, the complete method reduces nodes per recovered vessel from 1.957 to 1.298 and fragmented vessels from 25 to 11 on the south bank. Every vessel recovered by C050 remains recovered across all twelve sensitivity settings, and no extra vessel pair appears in the multi-vessel merge diagnostic. At the primary setting, the VLM additionally recovers one southern vessel. This result indicates that selected multi-view scene evidence can resolve much of the remaining physical fragmentation beyond geometric-semantic consolidation without increasing the observed multi-vessel merge evidence. The final representation retains the same 111/112 semantic agreement on the supported discrete objects.

The successful object consolidation is also observed qualitatively on P00 and P03 where multi-camera lifting increases the number of B0 and C050 fragments considerably compared to the mono-camera lifting performed on MUT.

\subsection{RQ3: Persistent Maritime Spatial Hierarchy}
\label{sec:results_hierarchy}

\begin{figure}[tbp]
    \centering
    \includegraphics[width=0.95\linewidth]{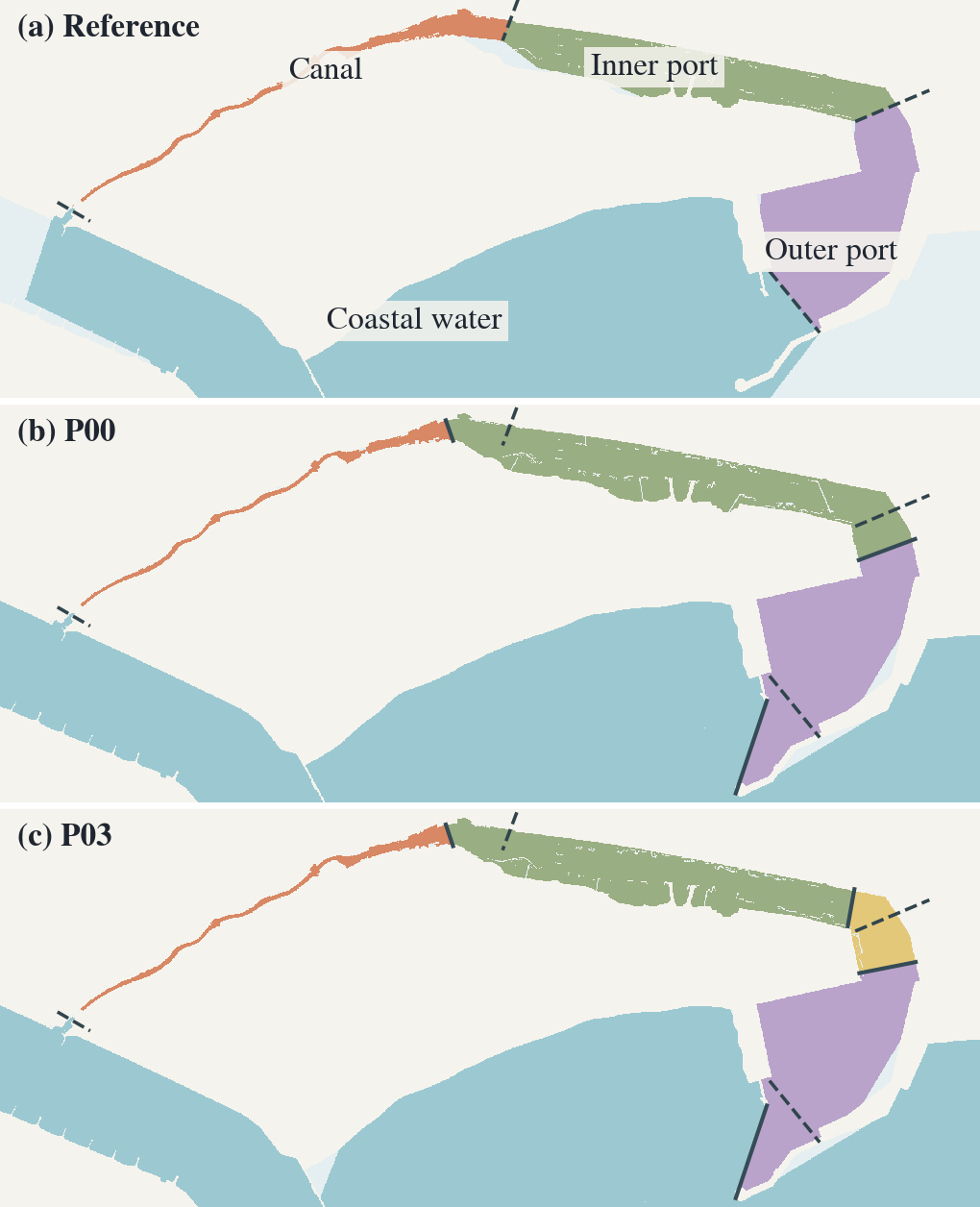}
    \caption{Macro regions compared. The first is defined by~\cite{chung2023pohang} while P00 and P03 are from our method. The regions are largely consistent, with one small extra region in P03.}
    \label{fig:macro}
\end{figure}

\paragraph{Macro regions}
The selected OSM plus onboard representation produces four regions in P00 and five in P03 with cross-traversal ARI 0.965 on common camera-water cells. The regions correspond closely with the four environmental areas identified independently in~\cite{chung2023pohang}, where a comparison to our method on P00 and P03 is shown in Fig.~\ref{fig:macro}. On assigned reference-covered water, equal-weighted mean IoU over the four one-to-one-matched reference areas is 0.812 for P00 and 0.810 for P03. Region assignments cover approximately 78.1\% and 78.0\% of the reference domain, respectively. The additional P03 region corresponds to a subdivision not present in the four-region reference. 

For comparison, onboard water/structure, the radar-augmented variant, and OSM alone give ARI 0.907, 0.623, and 1.000, respectively. OSM alone repeats by construction.

The combined representation therefore recovers a similar coarse spatial organization across the two traversals while remaining broadly consistent with~\cite{chung2023pohang}. The regions are not map-independent, but prior-assisted. The shared OSM topology provides the large-scale structure, while onboard water and structure observations provide traversal-specific observable real evidence.

\paragraph{Local places}
Local places are constructed independently in P00 and P03 before comparison. Across the ten matched geographical areas, we obtain 13 sample-associated places in P00 and 16 in P03. In total, 12 ordinary reciprocal correspondences are recovered, together with 3 partial alternatives and one type conflict among the partial relations. One P00 place and two P03 places remain unmatched.

\begin{figure}[tbp]
    \centering
    \includegraphics[width=0.87\linewidth]{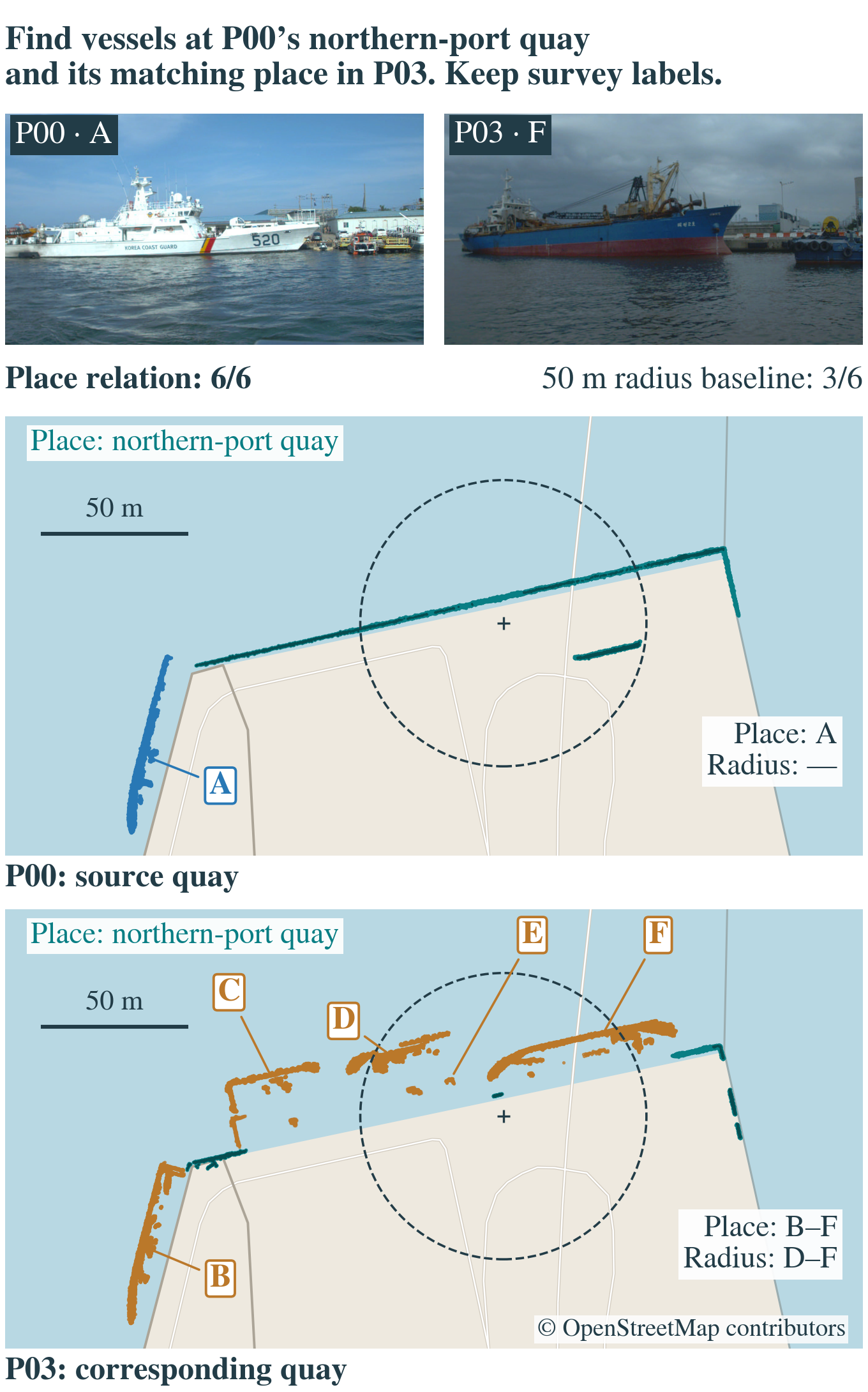}
    \caption{Place persistence and retrieval at the Northern port quay. The same fixed setting is matched despite different vessel occupancy and reconstructed geometry. Place relations retrieve $6/6$ reviewed vessel records across P00 and P03, compared with $3/6$ for the $50$\,m radius baseline centered at the P00 anchor location in both surveys.}
    \label{fig:persistent_context}
    \vspace{-15pt}
\end{figure}

The matched places include crossings, waterfronts, canopy-associated areas, among others. Reciprocal fixed-support overlap ranges from 0.341 to 0.770, reflecting both repeated structures and uncertain place extents. These results support repeatable local settings across traversals, while also showing that place extent remains less stable than the underlying fixed anchors. A persistent place example is shown in Fig.~\ref{fig:persistent_context}, where a place is recognized on both traversals despite P03 being severely blocked by vessels not present in P00. 

\subsection{RQ4: Contextual Retrieval}
\label{sec:results_retrieval}

We finally evaluate if our inferred hierarchy improves retrieval of entities and observations in a spatial context. Table~\ref{tab:retrieval} reports the primary Pohang evaluation. Flat entity-text retrieval performs poorly when queries require spatial context. Radius-based retrieval improves, but our hierarchy performs best on both single-survey and paired-survey queries.

\begin{table}[t]
\centering
\caption{Contextual retrieval on Pohang.}
\label{tab:retrieval}
\small
\setlength{\tabcolsep}{3pt}
\begin{tabular}{lcccc}
\hline
& \multicolumn{2}{c}{Single survey (/20)} &
\multicolumn{2}{c}{Paired (/4)} \\
Method & F1 & Exact & F1 & Exact \\
\hline
Flat entity text & 0.250 & 4 & 0.000 & 0 \\
Lexical radius & 0.387 & 4 & 0.150 & 0 \\
Location radius & 0.487 & 4 & 0.378 & 0 \\
Hierarchy & \textbf{0.787} & \textbf{11} & \textbf{0.810} & \textbf{2} \\
\hline
\end{tabular}
\vspace{-10pt}
\end{table}

An equivalent flat representation of the same relations gives the same answers. On the 16 context-conditioned single-survey queries, supplied-location radius F1 increases from 0.250 at $25$\,m to 0.579 at $100$\,m, but remains below the hierarchy. In one query (Fig.~\ref{fig:persistent_context}), for example, the hierarchy retrieves all six reviewed vessel records across the two surveys, despite very different spatial scenes, while the supplied-location radius retrieves three. The quay extends beyond the $50$\,m neighborhood around its P00 representative location.

The macro-context queries provide a test of the higher levels of the representation. Direct region queries are already handled well by geometric region assignment, with F1 0.967 for both geometry and hierarchy. For place-mediated macro queries, the hierarchy reaches F1 0.873 compared with 0.807 for a $50$\,m geometric support radius. Increasing the geometric radius to $100$\,m closes this gap. The hierarchy therefore provides an explicit way to propagate regional context through local-place relations, although sufficiently broad geometric lookup can recover the same information in these cases.

The hierarchy does improve contextual retrieval on the primary reviewed Pohang queries compared with entity-text and local radius-based representations, while the macro-context evaluation shows that some of the same context can be recovered using sufficiently broad geometric neighborhoods.
\section{DISCUSSION}
\label{sec:discussion_conclusion}

The results show that good semantics don't always lead to coherent 3D scene representation. In the maritime domain, large objects, limited viewing angles, and poor visibility can all be contributing factors to the high level of fragmentation observed with OpenGraph. While a lot of the perception and mapping pipeline can be maintained, physical identities and spatial abstractions needs to be adapted. 

We have seen that fixed infrastructure can support place correspondence even when reconstructed place extent and vessel occupancy differ considerably, as it can be seen in Fig.~\ref{fig:persistent_context}. Explicit place relations also help retrieval when a setting extends beyond a local metric neighborhood or must be related across traversals. The equivalent-flat control shows that this benefit comes from the inferred contextual relations rather than the graph data structure itself.

The resulting method can provide persistent context for different repeated maritime operations such as inspection, revisit planning and change detection. The current evaluation is limited to two datasets, prior-assisted macro regions and selected repeated-survey contexts, while dynamic, broader and longer-term evaluation remains future work.

\section{CONCLUSION}
\label{sec:conclusion}

Persistent scene memory is a prerequisite for autonomous maritime operations: a robot that repeatedly surveys the same waters should relate new observations to previously observed settings rather than treating each traversal independently. We show how an outdoor open-vocabulary 3DSG can be adapted into such a memory. Using OpenGraph as a representative baseline, our layer-wise transfer evaluation shows that open-vocabulary semantics transfer well to geometry-supported discrete maritime objects, while persistent physical identity and higher-level spatial organization do not: the mapped representation remains considerably fragmented, and the terrestrial organization assumptions don't provide the spatial structure required for maritime scene understanding.

We therefore keep the released perception and incremental mapping substrate and adapt its accumulated representation. Maritime-adapted consolidation and grounded multi-view vision-language reasoning form physical entities and local places, while maritime macro regions are built from geographic topology and onboard water and structure evidence. The complete adaptation substantially reduces physical-entity fragmentation while preserving vessel recovery, yields local places that correspond across independently constructed traversals despite different reconstruction and object occupancy, and improves paired-survey contextual retrieval over a local metric baseline. More broadly, the results show that 3DSGs can require domain adaptation even when the underlying semantic observations transfer well, providing persistent contextual memory for repeated maritime operations.


%

\section*{ACKNOWLEDGMENT}
OpenAI ChatGPT has assisted with brainstorming, critical discussion, and review, while Codex has supported software implementation, debugging, and experimental tooling. The authors performed and directed the research, developed and evaluated the ideas, validated any AI-assisted outputs, and retain full responsibility for the work.

\bibliographystyle{IEEEtran}
\bibliography{references}

@article{deng2024opengraph,
  title={Opengraph: Open-vocabulary hierarchical 3d graph representation in large-scale outdoor environments},
  author={Deng, Yinan and Wang, Jiahui and Zhao, Jingyu and Tian, Xinyu and Chen, Guangyan and Yang, Yi and Yue, Yufeng},
  journal={IEEE Robotics and Automation Letters},
  volume={9},
  pages={8402--8409},
  year={2024},
  publisher={IEEE}
}

@inproceedings{greve2024collaborative,
  title={Collaborative dynamic 3d scene graphs for automated driving},
  author={Greve, Elias and B{\"u}chner, Martin and V{\"o}disch, Niclas and Burgard, Wolfram and Valada, Abhinav},
  booktitle={IEEE International Conference on Robotics and Automation},
  pages={11118--11124},
  year={2024},
}

@article{lim2026hydra++,
  title={Hydra++: Real-Time Hierarchical 3D Scene Graph Construction With Object-Level Shape Estimation},
  author={Lim, Hyungtae and Hughes, Nathan and Yu, Xihang and Xu, Ruihan and Chang, Yun and Shi, Jingnan and Talak, Rajat and Carlone, Luca},
  journal={arXiv preprint arXiv:2607.09455},
  year={2026}
}

@article{hughes2022hydra,
  title={Hydra: A real-time spatial perception system for 3D scene graph construction and optimization},
  author={Hughes, Nathan and Chang, Yun and Carlone, Luca},
  journal={Robotics: Science and Systems},
  year={2022}
}

@article{schmid2024khronos,
  title={Khronos: A unified approach for spatio-temporal metric-semantic slam in dynamic environments},
  author={Schmid, Lukas and Abate, Marcus and Chang, Yun and Carlone, Luca},
  journal={Robotics: Science and Systems},
  year={2024}
}

@article{strader2024indoor,
  title={Indoor and outdoor 3D scene graph generation via language-enabled spatial ontologies},
  author={Strader, Jared and Hughes, Nathan and Chen, William and Speranzon, Alberto and Carlone, Luca},
  journal={IEEE Robotics and Automation Letters},
  volume={9},
  pages={4886--4893},
  year={2024},
  publisher={IEEE}
}

@inproceedings{berg2022using,
  title={Using language to generate state abstractions for long-range planning in outdoor environments},
  author={Berg, Matthew and Konidaris, George and Tellex, Stefanie},
  booktitle={International Conference on Robotics and Automation},
  pages={1888--1895},
  year={2022},
}

@inproceedings{nyffeler2025hierarchical,
  title={Hierarchical 3d scene graphs construction outdoors},
  author={Nyffeler, Jon and Tombari, Federico and Barath, Daniel},
  booktitle={IEEE/CVF International Conference on Computer Vision},
  pages={26817--26826},
  year={2025},
}

@inproceedings{steinke2025collaborative,
  title={Collaborative dynamic 3D scene graphs for open-vocabulary urban scene understanding},
  author={Steinke, Tim and B{\"u}chner, Martin and V{\"o}disch, Niclas and Valada, Abhinav},
  booktitle={IEEE/RSJ International Conference on Intelligent Robots and Systems},
  pages={6000--6007},
  year={2025},
}

@article{samuelson2025towards,
  title={Towards terrain-aware task-driven 3D scene graph generation in outdoor environments},
  author={Samuelson, Chad R and McLain, Timothy W and Mangelson, Joshua G},
  journal={arXiv preprint arXiv:2506.06562},
  year={2025}
}

@article{samuelson2025terra,
  title={Terra: Hierarchical terrain-aware 3D scene graph for task-agnostic outdoor mapping},
  author={Samuelson, Chad R and Austin, Abigail and Knoop, Seth and Romrell, Blake and Slade, Gabriel R and McLain, Timothy W and Mangelson, Joshua G},
  journal={arXiv preprint arXiv:2509.19579},
  year={2025}
}

@article{viswanathan2026xflie,
  title={xFLIE: Leveraging Actionable Hierarchical Scene Representation for Autonomous Semantic-Aware Inspection Missions},
  author={Viswanathan, Vignesh Kottayam and Saucedo, Mario Alberto Valdes and Satpute, Sumeet Gajanan and Kanellakis, Christoforos and Nikolakopoulos, George},
  journal={IEEE Transactions on Field Robotics},
  year={2026},
  publisher={IEEE}
}

@inproceedings{armeni20193d,
  title={3d scene graph: A structure for unified semantics, 3d space, and camera},
  author={Armeni, Iro and He, Zhi-Yang and Gwak, JunYoung and Zamir, Amir R and Fischer, Martin and Malik, Jitendra and Savarese, Silvio},
  booktitle={IEEE/CVF international conference on computer vision},
  pages={5664--5673},
  year={2019}
}

@article{rosinol20203d,
  title={3D dynamic scene graphs: Actionable spatial perception with places, objects, and humans},
  author={Rosinol, Antoni and Gupta, Arjun and Abate, Marcus and Shi, Jingnan and Carlone, Luca},
  journal={Robotics: Science and Systems},
  year={2020}
}

@article{rosinol2021kimera,
  title={Kimera: From SLAM to spatial perception with 3D dynamic scene graphs},
  author={Rosinol, Antoni and Violette, Andrew and Abate, Marcus and Hughes, Nathan and Chang, Yun and Shi, Jingnan and Gupta, Arjun and Carlone, Luca},
  journal={The International Journal of Robotics Research},
  volume={40},
  pages={1510--1546},
  year={2021},
  publisher={SAGE Publications Sage UK: London, England}
}

@inproceedings{gu2024conceptgraphs,
  title={ConceptGraphs: Open-Vocabulary {3D} Scene Graphs for Perception and Planning},
  author={Gu, Qiao and Kuwajerwala, Ali and Morin, Sacha and Jatavallabhula, Krishna Murthy and Sen, Bipasha and Agarwal, Aditya and Rivera, Corban and Paul, William and Ellis, Kirsty and Chellappa, Rama and Gan, Chuang and de Melo, Celso Miguel and Tenenbaum, Joshua B. and Torralba, Antonio and Shkurti, Florian and Paull, Liam},
  booktitle={IEEE International Conference on Robotics and Automation},
  pages={5021--5028},
  year={2024},
  doi={10.1109/ICRA57147.2024.10610243}
}

@inproceedings{koch2024open3dsg,
  title={{Open3DSG}: Open-Vocabulary {3D} Scene Graphs from Point Clouds with Queryable Objects and Open-Set Relationships},
  author={Koch, Sebastian and Vaskevicius, Narunas and Colosi, Mirco and Hermosilla, Pedro and Ropinski, Timo},
  booktitle={IEEE/CVF Conference on Computer Vision and Pattern Recognition},
  pages={14183--14193},
  year={2024},
  doi={10.1109/CVPR52733.2024.01345}
}

@inproceedings{werby2024hovsg,
  title={Hierarchical Open-Vocabulary {3D} Scene Graphs for Language-Grounded Robot Navigation},
  author={Werby, Abdelrhman and Huang, Chenguang and B{\"u}chner, Martin and Valada, Abhinav and Burgard, Wolfram},
  booktitle={Robotics: Science and Systems},
  year={2024},
  doi={10.15607/RSS.2024.XX.077}
}

@article{maggio2024clio,
  title={Clio: Real-Time Task-Driven Open-Set {3D} Scene Graphs},
  author={Maggio, Dominic and Chang, Yun and Hughes, Nathan and Trang, Matthew and Griffith, J. Daniel and Dougherty, Carlyn and Cristofalo, Eric and Schmid, Lukas and Carlone, Luca},
  journal={IEEE Robotics and Automation Letters},
  volume={9},
  pages={8921--8928},
  year={2024},
  doi={10.1109/LRA.2024.3451395}
}

@inproceedings{gassol2026relationship,
  title={Relationship-Aware Hierarchical {3D} Scene Graph},
  author={Gassol Puigjaner, Albert and Zacharia, Angelos and Alexis, Kostas},
  booktitle={IEEE International Conference on Robotics and Automation},
  year={2026}
}

@article{bovcon2021mods,
  title={{MODS}--A {USV}-Oriented Object Detection and Obstacle Segmentation Benchmark},
  author={Bovcon, Borja and Muhovi{\v{c}}, Jon and Vranac, Du{\v{s}}ko and Mozeti{\v{c}}, Dean and Per{\v{s}}, Janez and Kristan, Matej},
  journal={IEEE Transactions on Intelligent Transportation Systems},
  volume={23},
  pages={13403--13418},
  year={2022},
  doi={10.1109/TITS.2021.3124192}
}

@inproceedings{zust2023lars,
  title={{LaRS}: A Diverse Panoptic Maritime Obstacle Detection Dataset and Benchmark},
  author={{\v{Z}}ust, Lojze and Per{\v{s}}, Janez and Kristan, Matej},
  booktitle={IEEE/CVF International Conference on Computer Vision},
  pages={20304--20314},
  year={2023}
}

@article{benham2026coastline,
  title={The Coastline as a Structural Constraint: Harnessing Scene Geometry for Autonomous Surface Vessel Localization},
  author={Benham, Derek R. and Mangelson, Joshua G.},
  journal={arXiv preprint arXiv:2608.21276},
  year={2026},
  eprint={2608.21276},
  archivePrefix={arXiv},
  primaryClass={cs.RO},

}

@article{yang2024sgknet,
  title={{SGK-Net}: A Novel Navigation Scene Graph Generation Network},
  author={Yang, Wenbin and Qiu, Hao and Luo, Xiangfeng and Xie, Shaorong},
  journal={Sensors},
  volume={24},
  pages={4329},
  year={2024},
  doi={10.3390/s24134329}
}

@article{fagerhaug2025oceanscape,
  title={Oceanscape: A Graph-Based Framework for Autonomous Coastal Navigation},
  author={Fagerhaug, Eirik S. and Bye, Robin T. and Osen, Ottar L. and Hatledal, Lars Ivar},
  journal={Ocean Engineering},
  volume={320},
  pages={120230},
  year={2025},
  doi={10.1016/j.oceaneng.2024.120230}
}

@article{tonetti2026priorsg,
  title={{Prior-SG}: Task and Prior Driven Region Segmentation for Scene Graphs in Arbitrarily-Structured Environments},
  author={Tonetti, Giorgio and Kneip, Laurent and Gawel, Abel and Hutter, Marco},
  year={2026},
  eprint={2608.06170},
  archivePrefix={arXiv},
  primaryClass={cs.RO},
  journal={arXiv preprint arXiv:2606.06170},
}

@article{rotondi2026scenegraphs,
  title={{3D} Scene Graphs: Open Challenges and Future Directions},
  author={Rotondi, Dennis and Argenziano, Francesco and Koch, Sebastian and Hughes, Nathan and B{\"u}chner, Martin and Wald, Johanna and Rosenberger Schmid, Lukas and Nardi, Daniele and Valada, Abhinav and Paull, Liam and Tombari, Federico and Carlone, Luca and Arras, Kai O.},
  year={2026},
  eprint={2606.19383},
  archivePrefix={arXiv},
  primaryClass={cs.RO},
  journal={arXiv preprint arXiv:2606.19383}
}

@inproceedings{dalhaug2025near,
  title={Near-shore mapping for detection and tracking of vessels},
  author={Dalhaug, Nicholas and Stahl, Annette and Mester, Rudolf and Brekke, Edmund F{\o}rland},
  booktitle={International Conference on Information Fusion},
  pages={1--8},
  year={2025},
}

@inproceedings{bickici2026thinkgraphs,
  title     = {Think While You Map: Asynchronous Vision-Language Agents
               for Incremental {3D} Scene Graphs},
  author    = {Bickici, Deniz and Pabst, Michael and Mori, Shohei
               and Schmalstieg, Dieter},
  booktitle = {European Conference on Computer Vision},
  year      = {2026},
  eprint    = {2606.31471},
  archivePrefix = {arXiv},
}

@inproceedings{gorlo2026daaam,
  title     = {Describe Anything Anywhere At Any Moment},
  author    = {Gorlo, Nicolas and Schmid, Lukas and Carlone, Luca},
  booktitle = {IEEE/CVF Conference on Computer
               Vision and Pattern Recognition},
  pages     = {35002--35013},
  year      = {2026},
}

@article{hao2026ovipsg,
  title        = {{OVIP-SG}: Open-Vocabulary Instance-Preserving Scene Graphs
                  for Mapping and Retrieval of Small, Fine-Grained Objects},
  author       = {Hao, Tianjing and Lan, Haiyu and Li, Angsong and
                  Chen, Cheng and Li, Enyu and Yang, Jiarui and Su, Yuning
                  and Lin, Peiwen and Chuang, Wang},
  year         = {2026},
  eprint       = {2608.17633},
  archivePrefix = {arXiv},
  primaryClass = {cs.RO},
  journal={arXiv preprint arXiv:2608.17633},
}

@article{vanbaelen2022dynamic,
  title   = {Dynamic Semantic World Models and Increased Situational
             Awareness for Highly Automated Inland Waterway Transport},
  author  = {Van Baelen, Senne and Peeters, Gerben and Bruyninckx, Herman
             and Pilozzi, Paolo and Slaets, Peter},
  journal = {Frontiers in Robotics and AI},
  volume  = {8},
  pages   = {739062},
  year    = {2022},
  doi     = {10.3389/frobt.2021.739062}
}

@article{Brekke2022,
_doi = {10.1088/1742-6596/2311/1/012029},
_url = {https://dx.doi.org/10.1088/1742-6596/2311/1/012029},
year = {2022},
publisher = {IOP Publishing},
volume = {2311},
pages = {012029},
author = {Brekke, Edmund F and Eide, Egil and Eriksen, Bj{\o}rn-Olav H and Wilthil, Erik F and Breivik, Morten and Skjellaug, Even and Helgesen, {\O}ystein K and Lekkas, Anastasios M. and Martinsen, Andreas B and Thyri, Emil H. and Torben, Tobias and Veitch, Erik and Alsos, Ole A and Johansen, Tor Arne},
title = {{milliAmpere: An Autonomous Ferry Prototype}},
journal = {Journal of Physics: Conference Series},
abstract = {In this paper, we summarize the experiences with the autonomous passenger ferry development prototype milliAmpere, which has been used as a test platform in several research projects at the Norwegian University of Science and Technology (NTNU) since 2017. New algorithms for motion planning, motion control, collision avoidance, docking, multi-target tracking and localization have been developed and verified in full-scale experiments with milliAmpere. The infrastructure surrounding milliAmpere includes several sensor rigs supporting research on multi-sensor fusion and situational awareness, and a shore control lab which can be used to study the interaction between human operators and the autonomous ferry. Building upon the experiences with milliAmpere, the full-scale autonomous ferry milliAmpere2 was recently launched.}
}

@inproceedings{blanke2024greenhopper,
  title={{GreenHopper}: The Danish spearhead towards autonomous waterborne mobility},
  author={Blanke, Mogens and Hansen, P Nicholas and Dittmann, Kjeld and Enevoldsen, Thomas T and Dagdilelis, Dimitrios and Sch{\"o}ller, Frederik ETS and Plenge-Feidenhans, Martin K and Becktor, Jonathan and Papageorgiou, Dimitrios and Galeazzi, Roberto},
  booktitle={Journal of Physics: Conference Series},
  volume={2867},
  pages={012035},
  year={2024},
}

@article{akbar2021economic,
  title={An economic analysis of introducing autonomous ships in a short-sea liner shipping network},
  author={Akbar, Abeera and Aasen, Anna KA and Msakni, Mohamed Kais and Fagerholt, Kjetil and Lindstad, Elizabeth and Meisel, Frank},
  journal={International Transactions in Operational Research},
  volume={28},
  pages={1740--1764},
  year={2021},
  publisher={Wiley Online Library}
}

@article{dalhaug2026maritime,
  title={Maritime urban tracking dataset in harbor environment},
  author={Dalhaug, Nicholas and Nyg{\aa}rd, Trym Anthonsen and Hinostroza, Miguel and Flemmen, Henrik Dobbe and Martens, Emil and Stahl, Annette and Mester, Rudolf and Brekke, Edmund F{\o}rland},
  journal={Scientific Data},
  volume={13},
  pages={818},
  year={2026},
  publisher={Nature Publishing Group UK London}
}

@article{chung2023pohang,
  title={Pohang canal dataset: A multimodal maritime dataset for autonomous navigation in restricted waters},
  author={Chung, Dongha and Kim, Jonghwi and Lee, Changyu and Kim, Jinwhan},
  journal={The International Journal of Robotics Research},
  volume={42},
  pages={1104--1114},
  year={2023},
  publisher={SAGE Publications Sage UK: London, England}
}

@article{sawada_mapping_2023,
  title   = {Mapping and localization for autonomous ship using {LiDAR SLAM} on the sea},
  author  = {Sawada, Ryohei and Hirata, Koichi},
  journal = {Journal of Marine Science and Technology},
  volume  = {28},
  pages   = {410--421},
  year    = {2023},
  doi     = {10.1007/s00773-023-00931-y}
}

@article{engstrom_lidar-only_2022,
  title   = {A lidar-only {SLAM} algorithm for marine vessels and autonomous surface vehicles},
  author  = {Engstr{\"o}m, Artur and Geiseler, Domenic and Blanch, Krister and Benderius, Ola and Garc{\'i}a Daza, Iv{\'a}n},
  journal = {IFAC-PapersOnLine},
  volume  = {55},
  pages   = {229--234},
  year    = {2022},
  doi     = {10.1016/j.ifaco@article{christensen2026foundation,
  title={Foundation models on the bridge: Semantic hazard detection and safety maneuvers for maritime autonomy with vision-language models},
  author={Christensen, Kim Alexander and Tufte, Andreas Gudahl and Gusev, Alexey and Sinha, Rohan and Ganai, Milan and Alsos, Ole Andreas and Pavone, Marco and Steinert, Martin},
  journal={Ocean Engineering},
  volume={359},
  pages={124646},
  year={2026},
  publisher={Elsevier}
}
l.2022.10.436}
}

@article{christensen2026foundation,
  title={Foundation models on the bridge: Semantic hazard detection and safety maneuvers for maritime autonomy with vision-language models},
  author={Christensen, Kim Alexander and Tufte, Andreas Gudahl and Gusev, Alexey and Sinha, Rohan and Ganai, Milan and Alsos, Ole Andreas and Pavone, Marco and Steinert, Martin},
  journal={Ocean Engineering},
  volume={359},
  pages={124646},
  year={2026},
  publisher={Elsevier}
}

@misc{openai2026astra,
  author       = {{OpenAI}},
  title        = {{GPT-6 Astra}},
  howpublished = {OpenAI API documentation},
  url          = {https://developers.openai.com/api/docs/models/gpt-6-astra},
  note         = {Accessed: 2026-09-15}
}

@misc{openstreetmap,
  author = {{OpenStreetMap contributors}},
  title  = {{OpenStreetMap}},
  url    = {https://www.openstreetmap.org},
  note   = {Data available under the Open Database License}
}

\end{document}